\documentclass{article}
\PassOptionsToPackage{numbers}{natbib}

\usepackage[final]{neurips_2019}

\usepackage[utf8]{inputenc} 
\usepackage[T1]{fontenc}    
\usepackage{hyperref}       
\usepackage{url}            
\usepackage{booktabs}       
\usepackage{amsfonts}       
\usepackage{nicefrac}       
\usepackage{microtype}      

\newtheorem{thm}{Theorem}[section] 
\newtheorem{defn}[thm]{Definition} 

\usepackage{amsmath}
\usepackage{algorithm}
\usepackage{algpseudocode}
\usepackage[pdftex]{graphicx}
\usepackage{float}
\usepackage{wrapfig}

\title{Does Adam optimizer keep close to the optimal point?}

\author{%
  Kiwook Bae \thanks{Two authors contribute equally}\\
  KAIST\\
  \texttt{baekw92@kaist.ac.kr} \\
   \And
  Heechang Ryu \footnotemark[1]\\
  KAIST\\
  \texttt{rhc93@kaist.ac.kr} \\
   \And
   Hayong Shin \thanks{Corresponding author
    }\\
   KAIST \\
   \texttt{hyshin@kaist.ac.kr} \\
}

\begin{document}

\maketitle

\begin{abstract}
The adaptive optimizer for training neural networks has continually evolved to overcome the limitations of the previously proposed adaptive methods.
Recent studies have found the rare counterexamples that \textsc{Adam} cannot converge to the optimal point.
Those counterexamples reveal the distortion of \textsc{Adam} due to a small second momentum from a small gradient.
Unlike previous studies, we show \textsc{Adam} cannot keep closer to the optimal point for a general convex region when the effective learning rate exceeds the certain bound. Subsequently, we propose an algorithm that overcomes the \textsc{Adam}’s limitation and ensures that it can reach and stay at the optimal point region.
\end{abstract}

\section{Introduction}
First-order optimization algorithms have been practically used to train various machine learning models, including deep neural networks, with efficient computation and high performance. Stochastic gradient descent (SGD), one of the first-order optimization algorithms, is a method of iteratively learning the parameters of models using the negative gradient of the loss function with mini-batch extracted from data.

Subsequently, as variants of SGD, adaptive methods have been proposed that can use different learning rates for each parameter. \textsc{Adagrad} \citep{duchi2011adaptive}, the starter of adaptive methods, efficiently adjusted the learning rate for each parameter by dividing the learning rate by the square root of the sum of squares of the gradient vector of the parameter. Since then, adaptive methods have evolved to compensate for the weaknesses of previously proposed methods. \textsc{RMSprop} \citep{tieleman2012lecture} and \textsc{Adadelta} \citep{zeiler2012adadelta} have transformed the sum of \textsc{Adagrad}'s gradient vectors into an average, and \textsc{Adam} \citep{kingma2014adam} combined momentum with \textsc{RMSprop}. \textsc{Adam} has been still widely and practically used in training deep neural network because of its adaptive learning rate and excellent performance.

However, recent studies have pointed out the limitations of \textsc{Adam} optimizer \citep{reddi2019, zhou2018, luo2019}. They have shown counterexamples that \textsc{Adam} could not converge to the optimal point and suggested revised adaptive methods. \citet{reddi2019} have noted second momentum estimated by Exponential Moving Average (EMA) in \textsc{Adam} has a short term memory. If small gradients dominate the important large gradients, \textsc{Adam} moves with smaller step for important large gradients, and the optimizer could not converge to the optimal point. To overcome the limitation, they have proposed \textsc{AMSGrad}, a variant of \textsc{Adam}, which uses the max operator for second momentum estimate. The authors claim that the max operator keeps long term memory for the important large gradients. Additionally, \citet{zhou2018} have suggested another perspective of \textsc{Adam}'s shortcomings. If the correlation between gradient and second momentum estimate is broken in \textsc{Adam}, the optimizer could converge to the optimal point. They have suggested \textsc{AdaShift} optimizer which disconnects gradient and second momentum estimate. Finally, \citet{luo2019} have found out that extremely large learning rates by second momentum estimated from very small gradients can interfere with convergence. Therefore, \textsc{AdaBound} \citep{luo2019} clips the adaptive learning rates with scheduled upper$/$lower bound.




Recently, researchers are interested in how the optimizer, such as SGD, induces an optimal point in neural networks \citep{li17, kleinberg18}. The studies are based on the assumption that the neural network is a function which is convex near the optimal point $x^*$. In the studies, SGD could close to $x^*$ and stay with constant probability when the one point convexity with respect to $x^*$ holds.

In this paper, we show that the existing \textsc{Adam} cannot keep closer to or stay around the optimal point when the one point convexity with respect to the optimal point holds.
The counterexamples of previous studies \citep{reddi2019, zhou2018, luo2019} are the special cases of a stochastic convex optimization problem, which have an optimal point and a slightly worse suboptimal point. These counterexamples reveal the distortion of \textsc{Adam} due to a small second momentum from a small gradient. Unlike the counterexamples, we analyze the \textsc{Adam} in nonconvex problems with the assumption that one point convexity for the optimal point holds.
Furthermore, we derive the theoretical bounds of the second momentum, which prevents the optimizer from receding from an optimal point when they reach the optimal point region. From the bounds, we propose an algorithm that overcomes the limitation of \textsc{Adam} and ensures that it can reach and stay at the one point convex region with respect to the optimal point.







\section{Preliminaries}
We denote our objective function of $x\in\mathbb{R}^d$ as $f(x): \mathbb{R}^d \rightarrow \mathbb{R}$. Then, generic adaptive method could be written as follows \citep{reddi2019}:
\begin{center}
$x_{t+1}=x_t - \eta_t \frac{\hat{m_t}}{\hat{v_t}}$
\end{center}

where $\eta_t$ is a learning rate, and $\hat{m_t}$ is a typical momentum of gradients.
The key of adaptive method is to choose a function for second momentum $\hat{v_t}$. $\hat{v_t}$ is a nonnegative $d$-dimensional vector which adjusts learning rate for each dimension. \textsc{Adam} chooses an EMA of square of gradients for $\hat{v_t}$. \textsc{AMSGrad} adopts $\hat{v_t}=\max{\left \{ \hat{v_{t-1}},v_t \right \}}$. Therefore, the $\hat{v_t}$ in \textsc{AMSGrad} cannot be decreased. \textsc{AdaBound} clips $\hat{v_t}$ as $\hat{v_t}=clip\left ( v_t, \alpha_l(t), \alpha_u(t) \right )$ where $\alpha_l(t)$ and $\alpha_u(t)$ represent sequential lower and upper bounds.

\textbf{L-Smooth Function.}
We assume our objective \textit{f} is \textit{L-smooth} function. \textit{L-smooth} function is a typical and weak assumption for analysis of first-order optimization methods \citep{chen2018convergence, zaheer2018adaptive}. 

\begin{defn}
(L-smooth function). A function $f\in\mathbb{R}^d \rightarrow \mathbb{R}$ is L-smooth, if there exists a constant L such that,
\begin{center}
$\left \| \nabla f(x) - \nabla f(y) \right \|_2 \leq L\left \| x - y \right \|_2$, $\forall x,y \in \mathbb{R}^d $
\end{center}
\end{defn}

\textbf{One Point Convexity.}
There are studies \citep{li17,kleinberg18} for understanding how SGD induces an optimal point in neural networks. In the studies, they consider that the neural networks are complex nonconvex functions and assume a weak condition called \textit{one point convexity} in order to guarantee convergence.

\begin{defn}
(One point strongly convexity). A function f(x) is called $\delta$-one point strongly convex in domain $\mathbb{D}$ with respect to an optimal point $x^*$, $if \ \forall x \in \mathbb{D}$, $\langle-\nabla f(x),x^*-x\rangle > \delta \left \| x^* - x \right \|_2^2$
\end{defn}

This definition assumes that the neighborhood of the optimal point is convex. \citet{kleinberg18} have shown that if one point convexity with respect to the desired solution $x^*$ holds, SGD will get close to the $x^*$, and then stay around $x^*$ with constant probability. The researchers have also empirically observed that one point convex properties are satisfied at the convergence points of neural networks. Besides, according to Theorem 3.3 in \citep{kleinberg18}, if the learning rate is large in the one point convex region, even full gradient descent could not close to the optimal point and get out of the region.

In this study, we assume that \textsc{Adam} optimizer could arrive at a one point convex region of desired point $x^*$. After \textsc{Adam} arrives at the region, we show whether \textsc{Adam} keeps closer to $x^*$ or goes farther away from $x^*$.

\begin{algorithm}
\begin{algorithmic}[1]
\caption{\textsc{AdaFix}}
    \While {training}
    \State $g_t=\nabla f_t(x_t)$
    \State   $m_t=\beta_1 m_{t-1}+(1-\beta_1)g_t$
    \If{$\max_i g_{t,i} \geq L\eta$}
        
    \State $v_t=\beta_2 v_{t-1}+(1-\beta_2)g_t^2$
    \Else
    \State  $v_t=v_{t-1}$
    \EndIf
        
    \State  $x_{t+1}=x_t-\eta \frac{\hat{m_t}}{\hat{v_t}}$
        
    \State  $g_t'=\nabla f_t(x_{t+1})$
        
    \State  $l_t=\frac{\left \| g_t'-g_t\right \|_2}{\left \| x_{t+1}-x_t\right \|_2}$
        
    \State  $L = \max(L,l_t)$
    \EndWhile

\end{algorithmic}
\end{algorithm}

\section{Methods}

\begin{thm}
Let $\mathbb{D}$ is the domain of $\delta$-one point strongly convexity w.r.t. $x^*$.


\begin{center}
For $x_t \in \mathbb{D}$, if $\sqrt{\max_{i}{v_{t,i}}}\leq \left ( \frac{\sqrt{\delta^2\left \| x_{t} - x^* \right \|^2_2+\left \| \nabla f(x_t) \right \|^2_2}}{\left \| x_{t} - x^* \right \|^2_2} - \delta \right )
, then \left \| x_{t+1} - x^* \right \|^2_2 \geq \left \| x_{t} - x^* \right \|^2_2$.
\end{center}
where $v_{t,i}$ represents the $i$-th coordinate element in $v_t$.
\end{thm}

In theorem 3.1, $x_t$ is in the domain where $\delta$-one point convexity holds with respect to $x^*$. While $x_t$ gets closer to $x^*$, $\nabla f(x_t)$ would decreases. Subsequently, $v_t$ decreases until it is adjacent to the $\nabla f(x_t)$. Decreasing $v_t$ brings out gradual increasing of \textit{effective learning rate}, $\eta/\sqrt{v_t}$, which is the learning rate divided by the square root of $v_t$. In the theorem, the condition that $\sqrt{\max_{i}{v_{t,i}}}$ is smaller than the bound $(\frac{\sqrt{\delta^2\left \| x_{t} - x^* \right \|^2_2+\left \| \nabla f(x_t) \right \|^2_2}}{\left \| x_{t} - x^* \right \|^2_2} - \delta)$ means that the effective learning rates of entire dimensions exceed the bound. 
Thus, $x_t$ starts to recede from the optimal point. The receding continues until $\sqrt{\max_{i}{v_{t,i}}}$ is greater than the bound.

In the convex problem, the receding behavior from the optimal point is not a critical issue from a convergence perspective. $x_t$ could converge to the optimal point again because the function is convex on the entire region in the convex problem. On the other hand, in the nonconvex problem, there is a possibility that $x_t$ in \textsc{Adam} rushes out from the one point convex region, and it is hard to come back to the region again.
Such receding behavior has been pointed out in the study on SGD \citep{kleinberg18}. The study proves that even $x_t$ in full gradient descent algorithm is receding from the optimal point $x^*$ if the learning rate exceeds a certain bound.
Motivated from this study, we show it by an example that \textsc{Adam} is farther away from the optimal point due to the large effective learning rate and eventually it leaves the region.
\begin{wrapfigure}[14]{r}{0.4\textwidth}
  \vspace{-10pt}
  \centering
  \includegraphics[width=0.4\textwidth]{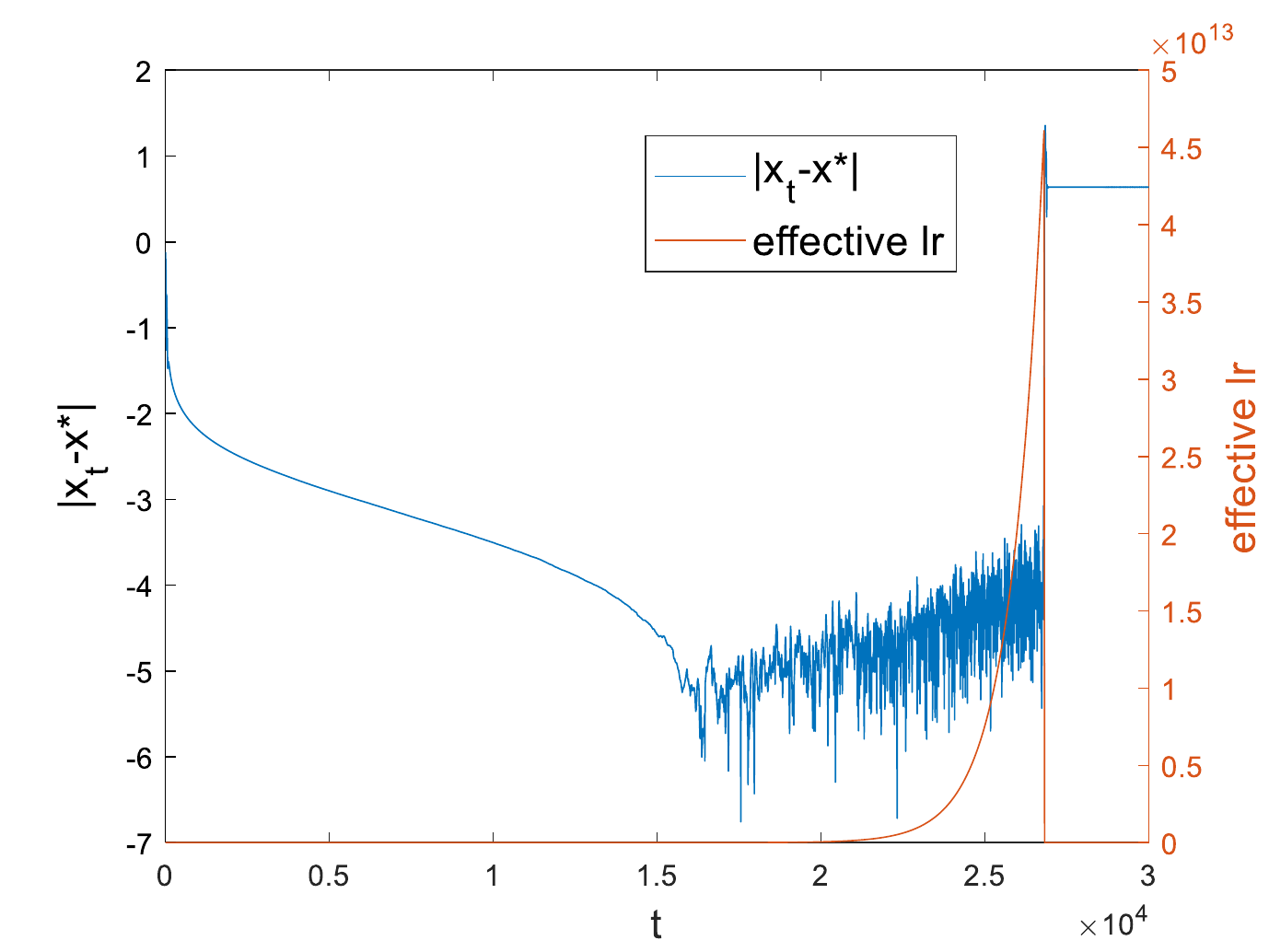}
  \vspace{-18pt}
  \caption{Experimental results on nonconvex function.}
  \label{fig:fig2}
\end{wrapfigure} 
We suggest a 2-dimensional nonconvex function, $f(x_1, x_2) = 1 - cos(x_1^2+x_2^2)$. The one point convexity holds with respect to $x^*=(0,0)$ in this example. 
We train $x$ with \textsc{Adam} to find $x^*$ to minimize the function from the initial point $x_0=(1.0, 0.3)$. \textsc{Adam} adopts $\eta=0.5, \beta_1=0.9$, and $\beta_2=0.999$.
Figure \ref{fig:fig2} shows that \textsc{Adam} approaches to the optimal point at the beginning of the training. However, $x_t$ moves farther away from the optimal point as the effective learning rate extremely increases during training. Subsequently, $x_t$ escapes from the one point convexity region.

To overcome this phenomenon, we propose a variant of \textsc{Adam}, referred to as \textsc{AdaFix}, which clips the second momentum once the entire elements of $g_t$ exceed the bound of theorem 3.1. However, the bound cannot be exactly calculated. We roughly estimate the bound with \textit{L-smooth} assumption. From definition 2.1 of \textit{L-smooth} function, $\left \| \nabla f(x_t) \right \|_2 \leq L\left \| x_t - x^* \right \|_2$ holds when $\nabla f(x^*)=0$. Then,
$\left ( \frac{\sqrt{\delta^2\left \| x_{t} - x^* \right \|^2_2+\left \| \nabla f(x_t) \right \|^2_2}}{\left \| x_{t} - x^* \right \|^2_2} - \delta \right ) \leq \eta (\sqrt{\delta^2+L^2}-\delta)$.
We also assume that \textit{L-smooth} constant L is much larger than $\delta$. This assumption implies the $\delta$-one point convex region would be flat while the neural network is fluctuating dramatically in the entire region. If $\delta \ll L$, then $\eta (\sqrt{\delta^2+L^2}-\delta) \cong L\eta$. 
We fix the entire second momentum once $\max_{i}{\left \{ g_{t,i} \right \}}$ equals to $L\eta$.

Our algorithm prevents the effective learning rate from getting too large and the point from escaping from the one point convexity region by fixing the second momentum after it is sufficiently small. We roughly estimate \textit{L-smooth} constant L in the bounds for the fixation along the training trajectory.
Previous study \citep{wood1996estimation} estimates L in the one dimensional functions. We apply the method to the multi-dimensional functions.

\textbf{Approximate L.}
From definition 2.1, $L \geq \frac{\left \| \nabla f(x) - \nabla f(y) \right \|_2}{\left \| x - y \right \|_2}$. We approximate L with $\max_{t}{l_t} = \frac{\left \| \nabla f_t(x_t) - \nabla f_t(x_{t+1}) \right \|_2}{\left \| x_t - x_{t+1} \right \|_2}$ \citep{wood1996estimation}.

        
        
        
        
        


\section{Experiments}

Compared to other algorithms, we validate our algorithm, \textsc{AdaFix}, for both data sets, MNIST and CIFAR-10, in Table \ref{table:table1} and \ref{table:table2}. SGDM in the tables stands for SGD with momentum. In the experiments in the tables, adaptive methods use a fixed learning rate, and SGDM uses a well-scheduled learning rate. The tables show that \textsc{AdaFix} has better or similar performances than other adaptive methods for various models and data sets, except for well-scheduled SGDM. In particular, \textsc{AdaFix} proposed to overcome \textsc{Adam}'s drawbacks outperforms the \textsc{Adam} in performance. 
\begin{table}[t]
  \caption{Results for MNIST}
  \label{table:table1}
  \centering
    \begin{tabular}{cc}
        \toprule
    
        \multicolumn{1}{c}{} & \multicolumn{1}{c}{Test Accuracy (\%)}\\
        \cmidrule(r){2-2}
        
        \multicolumn{1}{l}{Method}&\multicolumn{1}{c}{Two-layered MLP}\\
        \midrule
         \multicolumn{1}{l}{SGDM} & $98.72$\small $\pm0.02$ \\
         \cmidrule(r){1-2}
        \multicolumn{1}{l}{Adam}  & $98.70$\small $\pm0.03$\\
         \multicolumn{1}{l}{AMSGrad}& $98.72$\small $\pm0.07$ \\           
         \multicolumn{1}{l}{AdaBound}& $98.58$\small $\pm0.03$\\
         \multicolumn{1}{l}{AdaFix} & $\mathbf{98.75}$\small $\mathbf{\pm0.02}$\\
    
        \bottomrule
    \end{tabular}
\end{table}
\begin{table}[t!]
  \caption{Results for CIFAR-10 without and with augmentation}
  \label{table:table2}
  \centering
    \begin{tabular}{cccccc}
        \toprule
        \multicolumn{1}{c}{} & \multicolumn{5}{c}{Test Accuracy (\%)}\\
        \cmidrule(r){2-6}
        \multicolumn{1}{c}{} & \multicolumn{3}{c}{Without augmentation} & \multicolumn{2}{c}{With augmentation}\\
        \cmidrule(r){2-6}
        
        \multicolumn{1}{l}{Method}& \multicolumn{1}{c}{VGG11} & \multicolumn{1}{c}{ResNet34} & \multicolumn{1}{c}{DenseNet53}&
        \multicolumn{1}{c}{ResNet34} & \multicolumn{1}{c}{DenseNet53}\\
        \midrule
         \multicolumn{1}{l}{SGDM} & $84.66$\small $\pm0.28$ & $87.81$\small $\pm0.20$ & $89.46$\small $\pm0.03$& $\mathbf{93.55}$\small $\mathbf{\pm0.15}$ & $\mathbf{93.62}$\small $\mathbf{\pm0.13}$\\
         \cmidrule(r){1-6}
        \multicolumn{1}{l}{Adam}  &$84.82$\small $\pm0.18$  & $87.78$\small $\pm0.09$&$90.07$\small $\pm0.04$& $93.32$\small $\pm0.09$&$93.47$\small $\pm0.05$\\
         \multicolumn{1}{l}{AMSGrad}  & $\mathbf{85.91}$\small $\mathbf{\pm0.16}$ & $88.06$\small $\pm0.14$ &$90.00$\small $\pm0.17$& $93.38$\small $\pm0.13$ &$93.46$\small $\pm0.11$\\           
         \multicolumn{1}{l}{AdaBound}   & $83.23$\small $\pm0.21$ & $85.49$\small $\pm0.18$ & $86.53$\small $\pm0.25$& $91.56$\small $\pm0.11$ & $91.63$\small $\pm0.05$\\
         \multicolumn{1}{l}{AdaFix}    & $\mathbf{85.91}$\small $\mathbf{\pm0.18}$ &$\mathbf{88.08}$\small $\mathbf{\pm0.19}$ &$\mathbf{90.25}$\small $\mathbf{\pm0.19}$&$\mathbf{93.42}$\small $\mathbf{\pm0.18}$ &$\mathbf{93.51}$\small $\mathbf{\pm0.17}$\\
    
        \bottomrule
    \end{tabular}
\end{table}

\begin{figure}[b]
    
    \begin{minipage}[t]{.32\textwidth}
        \centering
        \includegraphics[width=\textwidth]{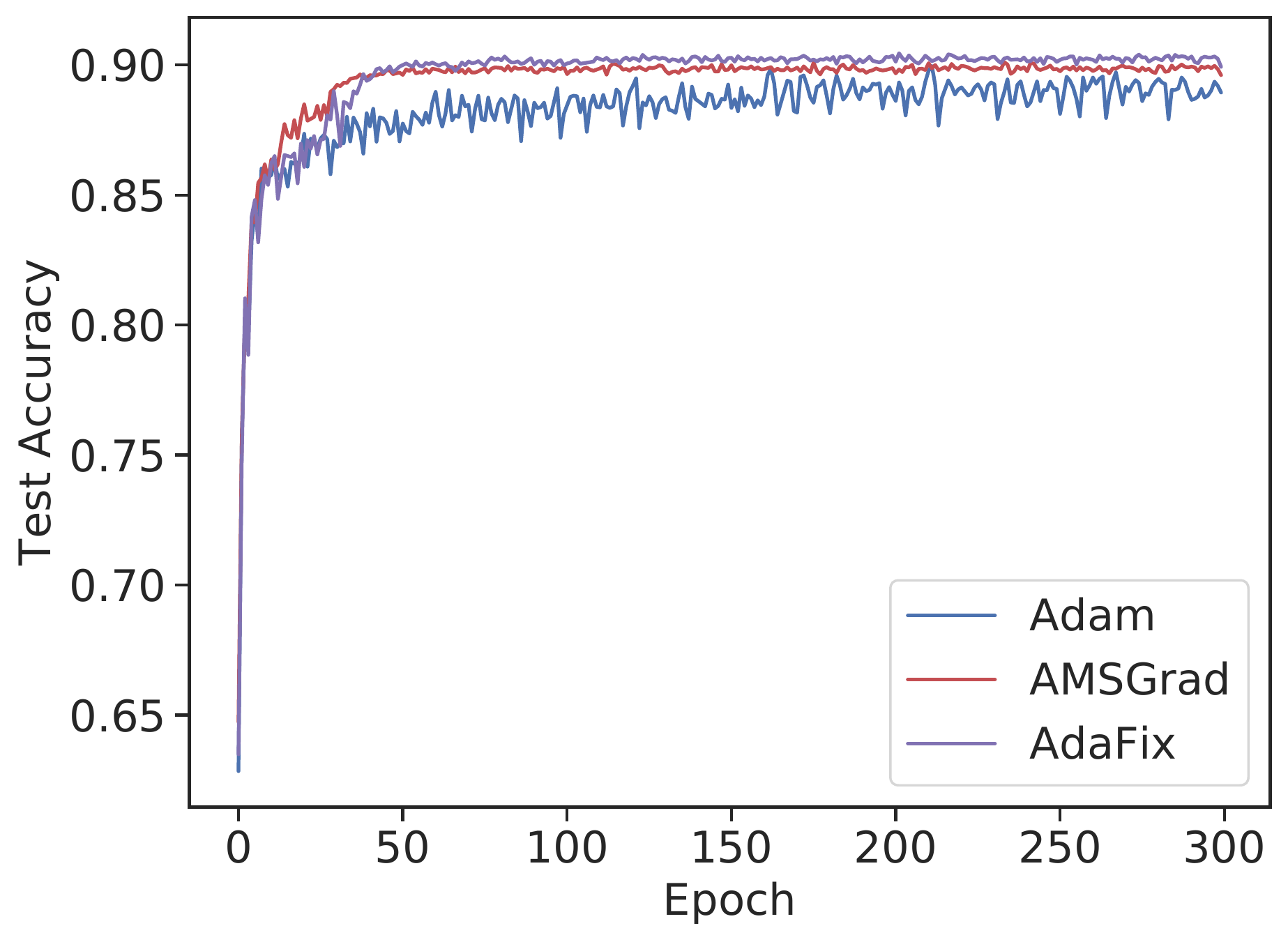}
    \end{minipage}
    \hfill
    \begin{minipage}[t]{.32\textwidth}
        \centering
            \includegraphics[width=\textwidth]{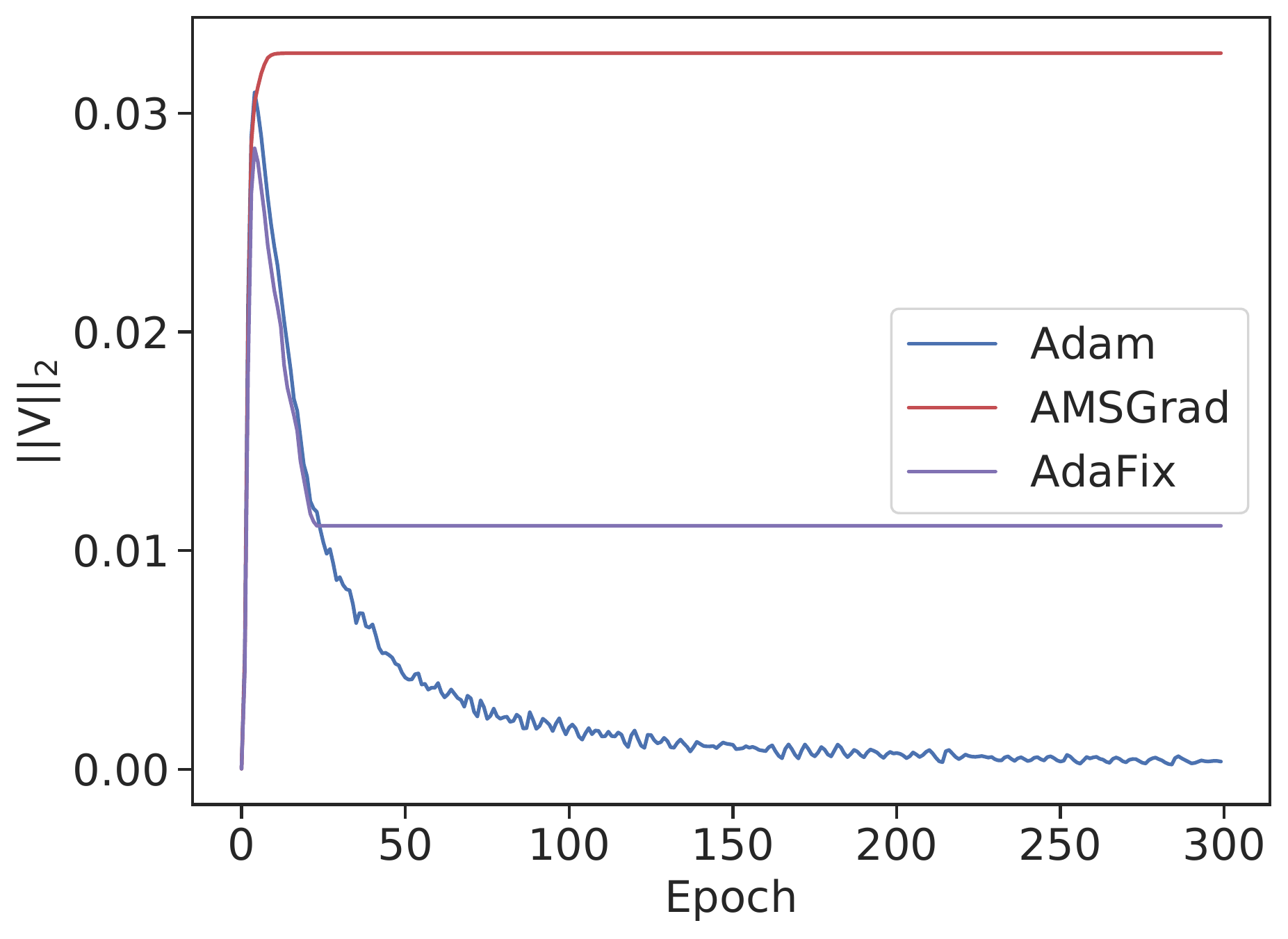}
    \end{minipage}  
    \hfill
    \begin{minipage}[t]{.32\textwidth}
        \centering
            \includegraphics[width=\textwidth]{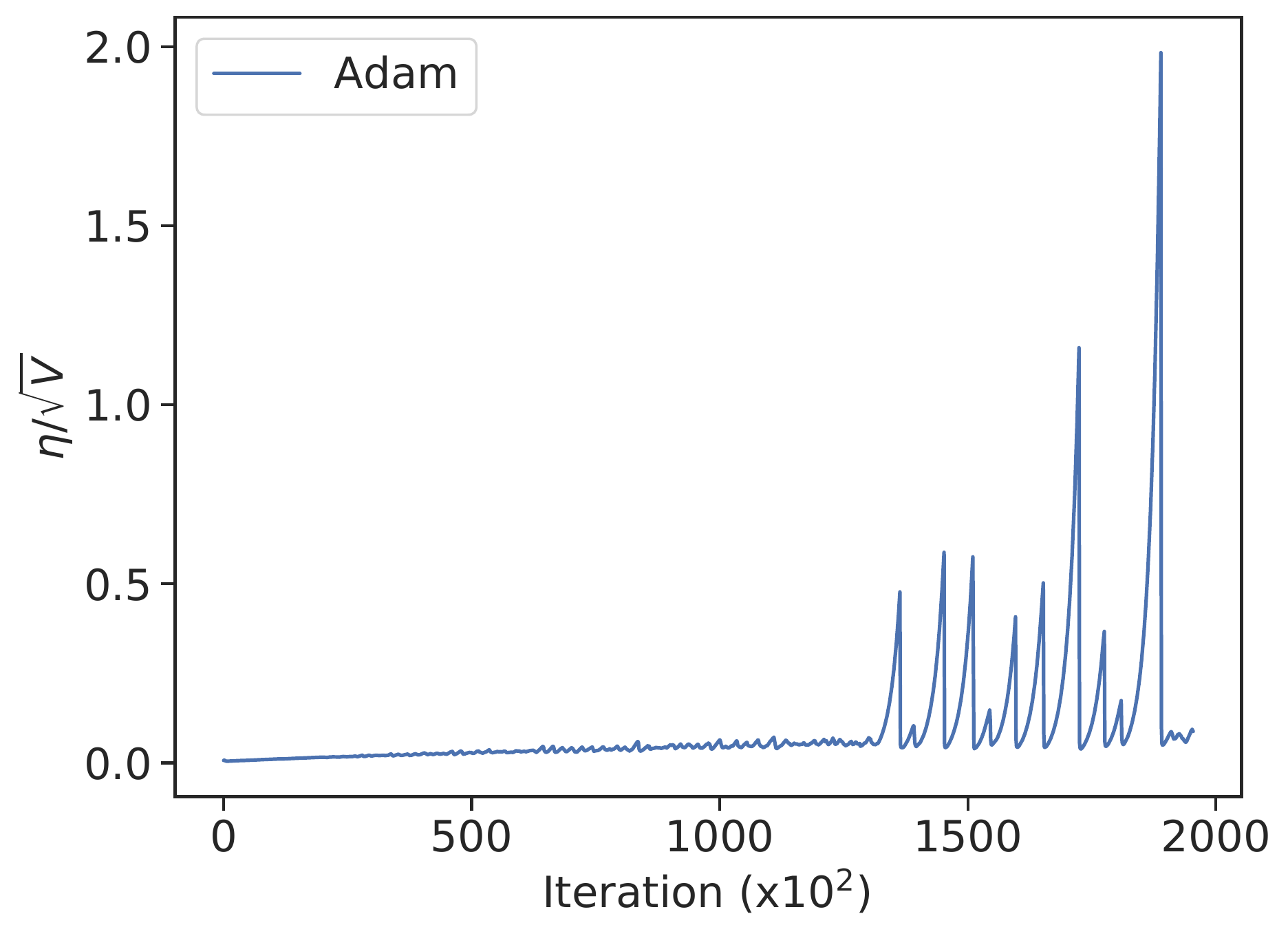}
    \end{minipage}  
    \vspace{-5pt}
    \caption{Results for CIFAR-10 to DenseNet53 without augmentation. Test accuracy, norm of $v$, and $\eta/\sqrt{v}$ are shown from left to right. }\label{fig:1}
\end{figure}

The success of \textsc{AdaFix} can be analyzed from Figure \ref{fig:1}. In that figure, \textsc{Adam}'s test accuracy seems very oscillating compared to \textsc{AMSGrad} and \textsc{AdaFix}. This phenomenon is explained from the norm of the average of the squares of the gradient vector, $||v||_2$, and effective learning rate, $\eta/\sqrt{v}$. In the figure, \textsc{Adam}'s $v$ becomes very small as it is learned, resulting in an explosive increase in the effective learning rate. Very large effective learning rates prevent learning from one point convex regions of optimal points. Meanwhile, \textsc{AMSGrad} prevents the effective learning rate from increasing by only taking increasing $v$, but this provides too little effective learning rate, and it is difficult to efficiently reflect the changing $v$ during training. \textsc{AdaFix} has better performances because it overcomes the limitation of \textsc{Adam} and \textsc{AMSGrad} by fixing $v$ in consideration of the characteristics of the function to be learned after fully reflecting the change of $v$.

    
        
    

\section{Conclusions}
In this paper, we have shown that the existing \textsc{Adam} cannot keep closer to or stay around the optimal point when the one point convexity with respect to the optimal point holds. Subsequently, we have proposed \textsc{AdaFix} that overcomes the \textsc{Adam}'s limitation and ensures that it can reach and stay at the one point convex region with respect to the optimal point.

\section{Acknowledgments}
This work was supported by the National Research Foundation of Korea (NRF) grant funded by the Korea government (MSIT). (No. 2017R1A2B4006290)

    

\bibliographystyle{plainnat}


\newpage

\section*{Supplementary Material}
\subsection*{Proof of Theorem 3.1}
$V_t$ is a diagonal matrix of second momentum.
$V_t= diag(v_t)+\epsilon I$

We set $v_{max} = \max_{i}{V_{t,ii}}$, and $v_{min} = \min_{i}{V_{t,ii}}$.

For convenience, we denote $A=V_t^{1/2}$.

Second momentum is non-negative. Therefore, following inequalities are true.

\begin{align*}
\left \| x_{t+1}-x^* \right \|_A^2= & \left ( x_{t+1} - x^* \right )^T A \left ( x_{t+1} - x^* \right ) \leq \left ( x_{t+1} - x^* \right )^T \sqrt{v_{max}}I\left ( x_{t+1} - x^* \right ) \\
& \leq \sqrt{v_{max}} \left \| x_{t+1} - x^* \right \|_2^2
\end{align*}

\begin{align*}
\left \| x_{t} - x^* \right \|_A^2 \geq \sqrt{v_{min}} \left \| x_{t} - x^* \right \|_2^2 
\end{align*}

The first equality follows from the \textsc{adam} update's rule. The second inequality follows from the $\delta$-one point strong convexity w.r.t. $x^*$.
\begin{align*}
\sqrt{v_{max}}\left \| x_{t+1} - x^* \right \|_2^2 & \geq \left \| x_{t+1}-x^* \right \|_A^2 = \left \| x_t - \eta A^{-1} \nabla f(x_t)-x^* \right \|_A^2  \\
& = \left \| x_{t} - x^* \right \|_A^2 - 2\eta \left ( A^{-1} \nabla f(x_t)\right )^T A \left( x_t - x^* \right) + \eta^2 \left \| A^{-1}\nabla f(x_t) \right \|_A^2 \\
& \geq \sqrt{v_{min}}\left \| x_t - x^* \right \|_2^2 - 2 \eta \delta \left \| x_t - x^* \right \|_2^2 + \eta^2 \left \| A^{-1/2}\nabla f(x_t) \right \|_2^2 \\
& \geq -2 \eta \delta \left \|x_t - x^* \right \|_2^2 + \eta^2 \left \| A^{-1/2}\nabla f(x_t) \right \|_2^2 \\
&\geq -2 \eta \delta \left \|x_t - x^* \right \|_2^2 + \frac{\eta^2}{\sqrt{v_{max}}} \left \| \nabla f(x_t) \right \|_2^2 \\
\end{align*}

If $\sqrt{v_{max}} \leq \eta \left ( \frac{\sqrt{\delta^2\left \| x_{t} - x^* \right \|^2_2+\left \| \nabla f(x_t) \right \|^2_2}}{\left \| x_{t} - x^* \right \|^2_2} - \delta \right )$, then 

\begin{align*}
-2 \eta \delta \left \|x_t - x^* \right \|_2^2 + \frac{\eta^2}{\sqrt{v_{max}}} \left \| \nabla f(x_t) \right \|_2^2 \geq \sqrt{v_{max}}\left \|x_t - x^* \right \|_2^2
\end{align*}

Therefore, we obtain the last inequailty.
\begin{align*}
\sqrt{v_{max}}\left \| x_{t+1} - x^* \right \|_2^2 \geq \sqrt{v_{max}}\left \|x_t - x^* \right \|_2^2
\end{align*}

\subsection*{Hyperparameters for Experiments}
For conducting the experiments in the main paper, the learning rate of all the adaptive optimizers is $0.001$, and the learning rate of SGDM starts at $0.1$ and decreases by $10$ times at $150$ and $250$ epochs. In addition, the momentum and second momentum parameters for EMA are $\beta_1=0.9$ and $\beta_2=0.999$, and the mini-batch size is $128$. Two-layered MLP with $2048$ units and ReLU for non-linear activation are used for the MNIST experiments. All experimental results are based on the results for $300$ epochs and five random seeds.

\end{document}